# Real-Time Video Highlights for Yahoo Esports

**Yale Song**
Yahoo Research
New York, USA
yalesong@yahoo-inc.com

## Abstract

Esports has gained global popularity in recent years and several companies have started offering live streaming videos of esports games and events. This creates opportunities to develop large scale video understanding systems for new product features and services. We present a technique for detecting highlights from live streaming videos of esports game matches. Most video games use pronounced visual effects to emphasize highlight moments; we use CNNs to learn convolution filters of those visual effects for detecting highlights. We propose a cascaded prediction approach that allows us to deal with several challenges arise in a production environment. We demonstrate our technique on our new dataset of three popular game titles, *Heroes of the Storm*, *League of Legends*, and *Dota 2*. Our technique achieves 18 FPS on a single CPU with an average precision of up to 83.18%. Part of our technique is currently deployed in production on Yahoo Esports.

## 1 Introduction

Esports is a form of competition on video games, where players compete with each other over prizes. The global esports market is growing fast, with an expected revenue of USD 463M and an audience of 256M people in 2016, which are 42.6% and 13.3% increases, respectively, compared to the previous year [9]. Several companies have recently launched websites dedicated to esports, e.g., *Twitch* and *YouTube Gaming*. Yahoo launched *Yahoo Esports* in March 2016 as the premier destination for delivering professional esports coverage across major games and events.

The flourishing amount of esports videos calls for an efficient way to index and share them over various channels. Video highlighting is an attractive method to achieve this with many potential applications. For example, one can share highlights on social media and use them to summarize game matches. In live broadcasting, highlights enable users to skim through the previously missed parts of a game, and allow companies to find ideal moments for programmatic ad placement.

Previous research on automatic video highlighting and summarization have focused on generic online videos [1, 12, 16], user generated content [8, 14], and sports videos [10, 2]. Recent approaches use deep neural networks to develop an end-to-end trainable system [3, 18]. The most related to our work is sports video highlighting [13]; early approaches include an analysis of audio signals [10] and text overlays [17]. Despite its practical importance, esports video highlighting has received relatively little attention from the research community.

In this paper, we present real-time video highlighting for Yahoo Esports. We observe that there are several pronounced visual effects in game highlights that set them apart from non-highlight game scenes, e.g., splash of lights with special moves, which suggests there is a great opportunity to use the latest computer vision techniques to solve the problem. There are, however, several challenges that arise in order to deploy our system in a production environment, such as the real-time performance requirement and an ability to deal with various types of scenes in esports broadcasting (e.g., interviews, commercials, etc.) that are not related to game scenes.

To address these challenges, we propose a cascaded prediction approach that uses two visual classifiers in a cascaded manner. The first determines if a frame contains game play and stops further analyzing it if not; the second determines if the frame contains a highlight. Crucial to the success of our system is a data set of over 300 hours of videos annotated with scene types and highlights at the frame level. We present our two-stage annotation scheme that allowed us to collect our data set quite efficiently.

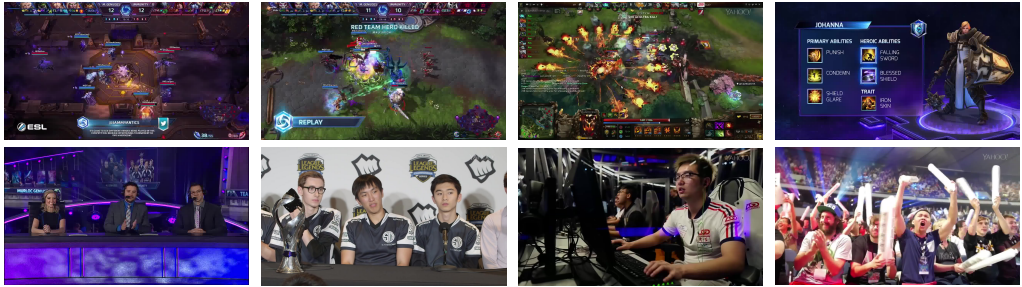

Figure 1: Esports live streaming videos contain various types of scenes. This increases the variability of the input space, creating challenges for esports highlighting. Images from left to right, top (game scenes): game play, game replay, game highlight, game character draft; and bottom (non-game scenes): commentator, interview, game player, and crowds. Our cascaded prediction approach effectively discards scenes unrelated to game, and detects highlights from game play scenes only.

## 2 Method

Figure 1 shows typical scenes that appear in esports live streaming videos, which can largely be categorized into game (top) and non-game (bottom) scenes. The former contains scenes that come from a computer game interface, including game play, game replay, and game character draft. The latter contains all other scenes, including commentators, player interviews, crowds, and commercials.

We are primarily interested in detecting highlights from game play scenes only. Our system, therefore, needs to distinguish game play scenes from game replay scenes; otherwise, we may have duplicate highlights, one from game play and another from game replay. The non-game scenes contain many sub-categories, making it difficult to specify clearly. But we believe doing so is unnecessary in our work because we care only about the game play scenes – all others simply need to be filtered out.

We formulate our problem as cascaded prediction with two visual classifiers: a scene type classifier and a highlight classifier. Each frame is first processed by the scene type classifier and categorized into one of four classes: game play, game replay, game character draft, and "others" that include non-game scenes. If a frame is a game play scene, it is subsequently processed by the highlight classifier and categorized into either highlight or non-highlight. We take a normalized confidence score of the highlight class as the highlight score, ranged between 0 and 1, and consider a frame a highlight by thresholding (in our experiments, we used a threshold of 0.5).

To achieve the real-time performance, we opted for a frame-based video analysis rather than sequence-based ones (e.g., 3D CNNs [15] or RNNs [4]). We sample and process every 5th frame of a video, and linearly interpolate the results to the original sampling rate. We can use any frame-based visual classifier as long as it provides good speed performance. We selected Convolutional Neural Networks (CNN) [7] as the base model for its empirical success on many visual classification tasks [11].

The entire system is implemented in C++ using the OpenCV and the CAFFE libraries. For the CNN we used the AlexNet [7] with batch normalization [5] after each layer. We trained each model from scratch using the ADAM optimizer [6], with a mini batch of 128 frames and for 100 epochs. The training was done using the CaffeOnSpark library on Yahoo grid infrastructure.

Our system is simple yet effective. The two classifiers have a clear separation of learning problems: scene type categorization and highlight detection. The system achieves 18 FPS on a single CPU machine, allowing us to process videos in real-time (processing every 5th frame of video requires a minimum of 6 FPS). Part of our technique is currently deployed in production on Yahoo Esports.

## 3 Data Collection

We collected a dataset of esports videos for three popular game titles: *Heroes of the Storm* (HotS), *League of Legends* (LoL), and *Dota2*. Our dataset contains roughly 100 hours of videos for each game title, with a total of about 300 hours (see Table 1). All videos are live recordings of major esports games and events and include scenes that appear in the real-world esports live broadcasting scenario, such as interviews, studio scenes, game replays, etc.

| Game Title | Non-game | Game | | | | Total |
|---|---|---|---|---|---|---|
| | | Non-highlight | High. lvl1 | High. lvl2 | High. lvl3 | |
| HotS | 31h49m28s | 50h00m00s | 06h20m03s | 02h57m55s | 00h38m46s | 91h46m13s |
| LoL | 37h18m26s | 63h23m59s | 06h42m11s | 03h05m38s | 00h31m55s | 111h02m11s |
| Dota2 | 19h56m13s | 79h55m17s | 11h06m57s | 03h44m44s | 00h21m08s | 115h04m21s |
| *Total* | 89h04m07s | 193h19m17s | 24h09m12s | 09h48m18s | 01h31m50s | 317h52m47s |

Table 1: The total duration of videos per scene type in our dataset. The *non-game* includes all kinds of scenes but game play, i.e., game replay, game character draft, and "others."

Annotating videos of more than 300 hours is undoubtedly a challenging task, especially when it involves subjective measurements such as finding highlights. We instituted a two-stage labeling scheme that is designed to reduce complexity in annotating video highlights.

**Scene type annotation**. First, we categorize each part of a video into one of four scene types: game play, game replay, game character draft, and "others" that include commentators, crowds, etc.

We employ a machine-in-the-loop approach to do this efficiently. For each game title, we begin by annotating from scratch a small batch of videos (about 10 hours). We then use it to train a scene type classifier (i.e., the CNN explained in the previous section), and use the trained model to predict labels for the next batch (about 20 hours). The predicted labels guide the second round of annotation, which involves correcting "mistakes" in the prediction results, rather than providing labels from scratch. Once the second batch is finished, we combine all annotated videos to train a new classifier, and use it to come up with predictions for the next batch (about 30 hours). We iterate this until we annotate all videos; each game typically takes about four rounds.

Because this stage involves objective judgment, we opted for having one expert annotator go through all the videos in our dataset; an esports editor from Yahoo Esports volunteered for this role. This helped us maintain consistency across videos and obtain high quality labels.

**Highlight annotation.** Next, we identify highlight moments from game play scenes. Unlike scene type labels, finding highlights from a video is a subjective task that can benefit from multiple measurements from different annotators. We therefore employ a crowdsourcing task in this stage.

We designed a web interface that allows annotators to find and identify interesting moments in a video. With it, one can adjust the video playback speed and skip parts that are not game play scenes, using the scene type labels obtained from the previous stage. After many iterations and considering feedback received from Yahoo Esports editors, we opted for using categorical labels to indicate different levels of highlights: level 0 (non-highlight), level 1 (cool), level 2 (wow), and level 3 (OMG). To ensure high quality labels, we had an influential figure in the esports community to personally reach out to the esports fans and enthusiasts, and recruit annotators who regularly watch live streamed esports videos. This allowed us to collect labels that contain semantic highlights (e.g., main character dies) rather than mere low-level visual highlights (e.g., splash of lights). The annotators were monetarily compensated for their efforts.

On average 4 annotators labeled each video (min:3, max:7, median:4). The inter-rater reliability in terms of the Cronbach's alpha was 0.7627. To further increase the quality of labels, for each video we chose three "best" annotators who maximally agreed with each other according to the Cronbach's alpha. This resulted in an increase in the Cronbach's alpha to 0.9225. We use an average of highlight scores from the best 3 annotators of each video as our final highlight label.

## 4 Evaluation

We treated each game title separately and performed three sets of experiments, one per game. For each game, we split our dataset so that 60% is used for training, 20% for validation, and 20% for test. We report our results in terms of average precision (AP) and recall at the frame-level.

**Models**. We evaluated six approaches, largely grouped into *Single* and *Cascade*. The former uses a single model to detect highlights, while the latter uses two models – a scene type classifier and a highlight detector – in a cascaded manner. All the classification models used the softmax function.

| Model | | HotS | | LoL | | Dota2 | |
|---|---|---|---|---|---|---|---|
| | | AP | Recall | AP | Recall | AP | Recall |
| Single | Random | 33.19 | 49.65 | 31.06 | 49.24 | 34.12 | 49.27 |
| | Binary | 62.31 | 75.14 | 38.33 | 43.93 | 57.86 | 61.40 |
| | Multiclass | 55.96 | 57.77 | 29.26 | 16.77 | 61.00 | **81.27** |
| Cascade | Random | 38.78 | 49.49 | 34.05 | 49.19 | 35.73 | 49.20 |
| | Regression | 51.72 | 19.57 | 26.67 | 9.83 | 50.00 | 44.08 |
| | **Binary (Ours)** | **83.18** | **86.29** | **59.66** | **56.76** | **67.90** | 77.22 |

Table 2: Evaluation results for each of the three game titles. Our approach outperforms all baselines in terms of AP and recall, except for recall on Dota2.

*Single-Random* produces random scores for all frames. *Single-Binary* uses a single binary classifier with one class representing game highlight and another representing all the other scene types. *Single-Multiclass* is a single classifier with 5 categories: game highlight, game play, game replay, game character draft, and all the others.

All cascade models shared one scene type classifier with 4 categories: game play, game replay, game character draft, and all the others. We evaluated four variants of highlight detector. *Cascade-Random* produces random scores for all frames categorized as game play. *Cascade-Regression* directly estimates the highlight score; we used a Euclidean loss and considered a frame a highlight if the score is above 1.0. *Cascade-Binary* is our model that detects highlights using a binary classifier.

**Results**. Table 2 shows our experimental results in terms of AP and recall, for each of the three game titles. Our approach (*Cascade-Binary*) consistently outperforms all other baselines in terms of AP and recall. We make several interesting observations.

We see that *Cascade-Random* consistently outperforms *Single-Random* in terms of AP. This shows the benefit of the cascaded prediction approach: Pre-filtering non-game scenes helps reduce making mistakes in highlight detection. It is also due to the fact that our scene type prediction model is very accurate, with above 99% AP and recall rates in all three games (it is a relatively easy task to discriminate game scenes from non-game scenes, because their pixel distributions are very different).

Comparing our approach against *Single-Binary* and *Single-Multiclass* shows the benefit of our cascaded prediction approach. All three models perform classification to detect highlights; the only difference is that we use an additional scene type classifier to filter out non-game scenes. This greatly improves AP, with a small trade off in speed performance (a single model achieves 28 FPS).

Comparing our approach against *Cascade-Regression* shows that it is better to perform classification on binarized highlight scores rather than directly estimating it. The regression approach performs particularly worse in recall, missing most of the true highlight scenes. This is due in part to the heavily imbalanced sample distribution; the non-highlight sample size is larger than highlight level 3 by more than 100 times (see Table 1).

## 5 Conclusion

We presented a cascaded prediction approach to detect highlights from esports live streaming videos, and showed the benefit of our approach on three popular game titles with over 300 hours of videos. One of our primary goals in this work was to develop a system that is fast enough to be used in a production environment. We achieved the real-time performance via simple design of the system.

There are, of course, many research areas we would like to explore further. One is transfer learning from one game to another. While there exists many game titles, there are only a few esports genres, such as real-time strategy, fighting, first-person shooter, and multi-player online battle arena (our three game titles belong to this genre). Games in each genre share certain similarities in scene layouts and visual effects, with subtle differences between games. This makes our data set an interesting test bed for evaluating transfer learning techniques. Another is multimodal video processing. This work used visual signals only, but audio and text signals also provide valuable information; commentators scream during highlight moments, and game interfaces show text overlay during important moments (e.g., *"triple kill"*). We look forward to using our large dataset to explore those areas in the future.

## Acknowledgement

We thank Jordi Vallmitjana and the Yahoo Esports editorial team for their support on data collection, and the Yahoo Esports engineering team for their support on production deployment of our system.